\documentclass[letterpaper, 10 pt, conference]{ieeeconf}
\IEEEoverridecommandlockouts % for \thanks
\usepackage{graphicx} % for \includegraphics
\usepackage{booktabs}
\usepackage[colorlinks, citecolor=red]{hyperref}
\usepackage{cite}
\usepackage{float}
\usepackage{comment}
\graphicspath{{./pdf/}} % load figure within subfolder ./pdf/
\DeclareGraphicsExtensions{.pdf}
\usepackage{algorithm}
\usepackage{algorithmic}

\renewcommand{\algorithmiccomment}[1]{\bgroup\hfill//~#1\egroup}
\usepackage[export]{adjustbox} % for \adjincludegraphics

\usepackage[cmex10]{amsmath} % assumes amsmath package installed

\usepackage[short]{optidef} % for argmini environment

\usepackage{bm} % for \bm

\usepackage{epsfig} % for postscript graphics files

\usepackage{times} % assumes new font selection scheme installed

\usepackage{color} % for \color

\usepackage{amssymb}  % assumes amsmath package installed
\usepackage{multirow}

\newcommand{\mytilde}{\raise.17ex\hbox{$\scriptstyle\mathtt{‌​\sim}$}}

\title{\LARGE \bf
	One to Many: Adaptive Instrument Segmentation via Meta Learning and Dynamic Online Adaptation in Robotic Surgical Video
}

\author{Zixu Zhao, Yueming Jin, Bo Lu, Chi-Fai Ng, Qi Dou, Yun-Hui Liu, and Pheng-Ann Heng% <-this % stops a space
	\thanks{This work was supported by Key-Area Research and Development Program of Guangdong Province, China (2020B010165004), Hong Kong RGC TRS Project No.T42–409/18-R, National Natural Science Foundation of China with Project No.U1813204,
 and CUHK Shun Hing Institute of Advanced Engineering (project MMT-p5–20). }
	\thanks{Z. Zhao, Y. Jin, Q. Dou and P. A. Heng are with the Department of Computer Science and Engineering, The Chinese University of Hong Kong, Hong Kong. \emph{Corresponding author: Yueming Jin (ymjin@cse.cuhk.edu.hk) }}
	\thanks{B. Lu, Q. Dou, and Y.H. Liu are with the T stone Robotics Institute, The Department of Mechanical and Automation Engineering, The Chinese University of Hong Kong, Hong Kong. }
	\thanks{P. A. Heng is also with Guangdong-Hong Kong-Macao Joint Laboratory of Human-Machine Intelligence-Synergy Systems, Shenzhen Institutes of Advanced Technology, Chinese Academy of Sciences, China}
	\thanks{C. F. Ng is with the Department of Surgery, The Chinese University of Hong Kong, Hong Kong}
}

\begin{document}

\maketitle
\thispagestyle{empty}
\pagestyle{empty}

\begin{abstract}

Surgical instrument segmentation in robot-assisted surgery (RAS) - especially that using learning-based models - relies on the assumption that training and testing videos are sampled from the same domain.
However, it is impractical and expensive to collect and annotate sufficient data from every new domain. To greatly increase the label efficiency, we explore a new problem, i.e., adaptive instrument segmentation, which is to effectively adapt \textit{one} source model to new robotic surgical videos from \textit{multiple} target domains, only given the annotated instruments in the first frame.
We propose  \textit{MDAL}, a meta-learning based dynamic online adaptive learning scheme with a two-stage framework to fast adapt the model parameters on the first frame and partial subsequent frames while predicting the results. MDAL learns the general knowledge of  instruments and the fast adaptation ability through the video-specific meta-learning paradigm. The added gradient gate excludes the noisy supervision from pseudo masks for dynamic online adaptation on target videos.
We demonstrate empirically that MDAL outperforms other state-of-the-art methods on two datasets (including a real-world RAS dataset). The promising performance on ex-vivo scenes also benefits the downstream tasks such as robot-assisted suturing and  camera control.
%It also achieves satisfying qualitative results on two ex-vivo robotic surgical videos to densely track the target instruments or tool tips.

\end{abstract}

%======================================================
\section{INTRODUCTION}

Robot-assisted surgery (RAS) has revolutionized the minimally invasive surgery by facilitating surgeons to perform complex and precise manipulations.
%Robot-assisted minimally invasive surgery can greatly improve the surgeon performance and patient safety with dexterous articulated instrument.
%For example, the dexterous articulated instruments in \textit{da Vinci} telerobotic surgical system can provide  surgeons with better control.
Intelligently understanding the robotic instruments by pixel-wise semantic segmentation is highly desired for promoting the cognitive assistance to surgeons.
%To further extend their perception with the endoscopic view, a pixel-wise segmentation of each frame in a surgical robotic video is currently required.
Precise instrument segmentation serves as a building block for tool pose estimation~\cite{allan20183}, tool tracking and control~\cite{du2019patch}, which are crucial for surgical robot navigation~\cite{allan20192017, gao2021future}.
%supporting more advanced downstream capabilities such as
Further providing the tool position by instrument segmentation contributes to automatic camera optimal view control, in a way of reducing the manual movement of the endoscopic camera~\cite{islam2020st}, hence benefiting robotic autonomous operation. %tracking
%The segmentation results can be used to generate early warning of potential deviations and anomalies.
%Such function benefits to achieve robotic autonomous operation.
The semantic segmentation also enables various automatic post-operative capabilities, such as surgeon skill evaluation, surgical report documentation, and surgeon educational training~\cite{reiley2009task,zia2018surgical}.

\begin{figure}[t]
	\centering
	\vspace{-1mm}
	\includegraphics[width=0.45\textwidth]{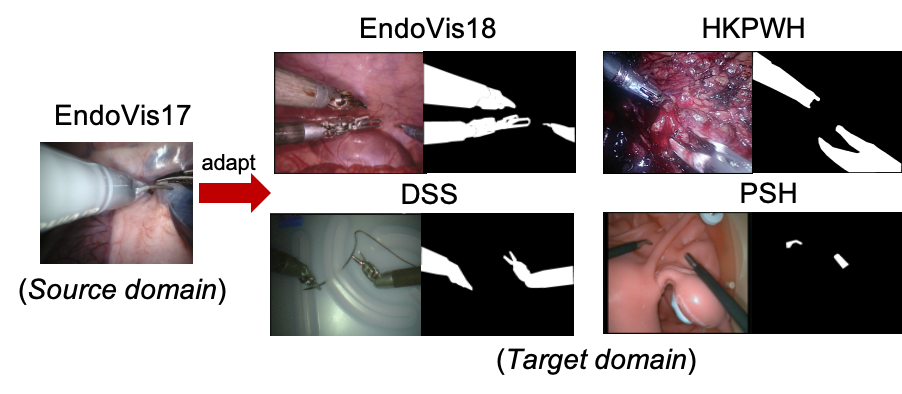}
	\caption{Typical endoscopic images from EndoVis17 dataset (source), EndoVis18, HKPWH, DSS, and PSH datasets (target). The segmentation targets vary from multiple instruments (EndoVis18, HKPWH, DSS) to tool tips (PSH).
	The distribution shift between two domains cannot be ignored.}
	\label{intro}
	\vspace{-3mm}
\end{figure}
Instrument segmentation from robotic surgical videos is challenging, due to the complicated scene,  instrument motion blur, and visual occlusion by blood or smoke.
To tackle these challenges, deep learning based methods  have been proposed and achieved promising segmentation results.
In particular, fully-supervised methods are dominant, but rely on large amounts of annotations for training~\cite{ yu2019holistically, islam2019real,islam2020ap,ni2020attention}.
For label-effective training, some semi-supervised, self-supervised, and domain adaptation methods are explored to generate  pseudo masks for unlabeled training data, respectively using the cycle consistency properties in robotic videos~\cite{jin2019incorporating, zhao2020learning}, robot kinematic model~\cite{da2019self,colleoni2020synthetic}, and simulated surgical scene~\cite{pfeiffer2019generating}.
However, most studies assume that testing data should be drawn from the same distribution as the training data, i.e., one RAS dataset for model training and testing.

In real-world RAS, videos from different clinical sources can differ significantly in data distribution (Fig.~\ref{intro}), which is mainly caused by different recording systems and imaging protocols.
Even from the same clinical site, surgical scene of each procedure may present various appearances caused by patient cohorts, unexpected deviation of blooding or surgeon operative skills.
Directly deploying the existing models would suffer from dramatic performance degradation once encountering the distribution shift issue between the training and testing data.
It is arguably impossible to build a dataset that includes all kinds of data distributions and train a universal model on it.
Also, it is expensive and sometimes impossible to fine-tune the pre-trained model for each clinical site or newly collected surgical datasets, considering the scarce nature of surgical data and annotations.

Unlike the domain adaptation scenario~\cite{sahu2020endo, chen2019blending, qian2019domain} where one specific model is trained for one target domain, given a deep-trained model, we expect that it can adapt to videos from one or many target domains with the low-data regime.
%Ideally, we expect a deep-trained model to has the capability to quickly adapt to new videos from unseen distribution with low-data regime.
In this paper, we focus on a new and crucial topic, \textit{adaptive  instrument segmentation}, which is to train a model from  easily-accessible surgical robotic videos (source domain), so that given a new video with only the first frame annotation available (target domain), we can fast adapt the model with few steps of gradient descent to consistently segment the instruments across frames, a.k.a dense instruments tracking.
The learned model can be applicable to the upcoming robotic operations, as long as we collect one initial scene and perform the labelling to specify the segmentation target.

One vanilla solution to this problem is  fine-tuning the pre-trained model on the first frame annotation in the target video so that it can adapt to the new distribution.  Unfortunately, the model requires hundreds of fine-tuning iterations to learn sufficient representation from the first frame, especially when we apply data augmentations.
%This is detrimental to the model's real-world RAS performance.
Model-agnostic meta-learning (MAML)~\cite{finn2017model} is a recent method for fast model adaptation on new domains or categories.
It provides a paradigm to learn the general representations from a bunch of similar tasks, which can be quickly adapted to the new task with very few iterations.
%~\cite{sun2019meta,jamal2019task} introduce image-level meta-learning schemes for few-shot learning the new class.
%~\cite{zhang2019online, tonioni2019learning} propose unsupervised methods for fast video adaptation in a simulated-to-real scenario on depth estimation task. ~\cite{xiao2019online} extends the meta-learning framework to semi-supervised video object segmentation by learning the common structure of natural objects.
MAML has been widely used in few-shot learning~\cite{sun2019meta,jamal2019task, garcia2017few}, and domain adaptation~\cite{chen2019blending, qian2019domain}.
Many of them are proposed to solve high-level vision tasks such as classification~\cite{sun2019meta,jamal2019task}, estimation~\cite{zhang2019online, tonioni2019learning, yu2020foal}, and segmentation~\cite{xiao2019online}. Such works inspire us to build a meta-learning based framework for this new problem, which teaches the model with fast adaptation capability.

\textbf{Contributions:} We propose \textit{MDAL}, a meta-learning based dynamic online adaptive learning scheme for adaptive instrument segmentation in robotic surgical videos. MDAL consists of an offline meta-training stage and a dynamic online adaptation stage. Firstly, MDAL trains the model with a meta-learner to  obtain the  fast adaptation capability on the new videos.
A video-level task space is introduced into the meta-learning  framework so that the model leans to capture transferable internal representations of source instruments. In the second stage, MDAL tests the new videos in the target domain while online adapting the model parameters across frames with the fast adaptation ability. Given the first frame annotation, the model meta-adapts on the target instruments with very few iterations. To address the appearance variation and fast instrument motion problems in robotic surgical videos,  MDAL further
dynamically  adapts the model on  partial subsequent frames via a noise-aware online adaptation strategy that monitors the perfection of induced pseudo masks, in which each adaptation step only requires one gradient descent. Our main contributions include:
\begin{itemize}

\item Taking the first step to investigate the crucial yet challenging problem for robotic surgical videos, i.e., \textit{adaptive instrument segmentation} (one-to-many adaptations). Improving the label efficiency with the only-first-frame-labelling rule in new domains.
%Building a universal model that fast adapt well to new videos from different distributions by providing only the annotation of the first frame, which greatly improves the data efficiency.

\item Designing a video-specific task space for meta-learning that enforces the model to capture the common knowledge of instruments shared across multiple video tasks.

\item Online adapting the model on testing frames via a noise-aware gradient gate that filters out the imperfect supervision and selectively updates the parameters. Improving both the adaptation performance and efficiency against continuous online adaptation.

\item Achieving an average instrument segmentation results of 75.5\% IoU and 84.9\% Dice on new domains.

\end{itemize}

MDAL was trained with a publicly available EndoVis17~\cite{allan20192017} dataset,
and extensively evaluated on a more complex EndoVis18~\cite{allan20202018} dataset containing porcine procedures, and a newly collected HKPWH dataset containing real-world prostatectomy. MDAL also achieves satisfying dense tracking results on ex-vivo scenes, including dVRK-based suturing videos (DSS) and phantom-based simulated hysterectomy videos (PSH), demonstrating the great potential of our method for crucial downstream tasks in RAS such as robot-assisted suturing and autonomous camera control.
%which can be directly applied to the downstream tasks including  suture grasping and autonomous camera control.

\begin{figure*}[t]
	\centering
	\includegraphics[width=0.9\textwidth]{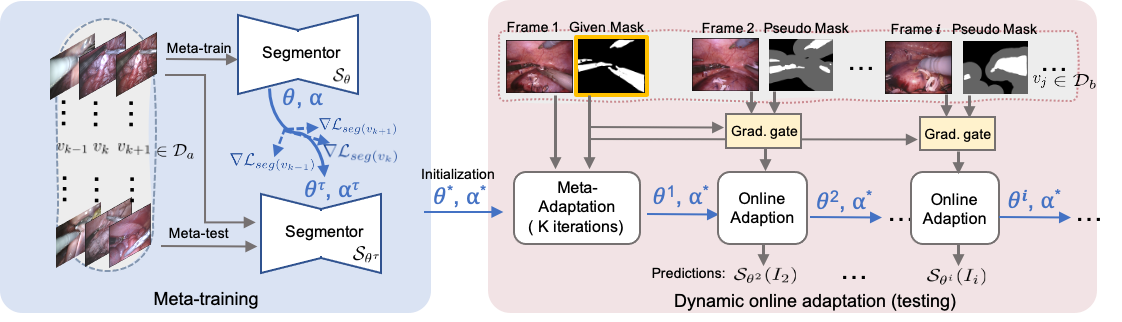}
	\vspace{-3mm}
	\caption{Overview of proposed MDAL. It contains two stages (i) meta-training: learning a meta-learner ($\theta, \alpha$) that provides the fast adaptive  parameters $\theta^{*}$ and step size $\alpha^{*}$ for the initialization of stage (ii); (ii) dynamic online adaptation: meta-adapting the model parameters  $\theta^{*}$ on the first frame mask with very few iterations $K$, and selectively online adapting  $\theta^{1}$ on the remaining frames via the pseudo masks and gradient gate.}
	\label{method}
	\vspace{-3mm}
\end{figure*}

\section{METHODS}
\label{METHODS}

The proposed MDAL adapts the segmentation model to target robotic surgical videos from different domains. As shown in Fig.~\ref{method}, MDAL involves two stages:
(1) \emph{meta-training} the model with a  meta-learner that captures the general knowledge from the task-specific source domain;
(2) \emph{online adapting} the model on the target robotic surgical videos in a dynamic manner.
The following subsections describe the problem setup, meta-learning scheme for training, and online adaptation strategy for testing.

\subsection{Problem Setup}
Adaptive instrument segmentation only requires the first frame  mask rather than amounts of annotations  from the target domain, compared to the fully-~\cite{yu2019holistically} and semi-~\cite{zhao2020learning} supervised setting. More practical than the domain adaptation scenario~\cite{qian2019domain}, it adapts one model to multiple domains without training multiple different models.
%Compared with the domain adaptation setting~\cite{sahu2020endo, chen2019blending, qian2019domain}, \textit{adaptive instrument segmentation} only requires one ground-truth mask rather than amounts of annotations from the target distributions. More practically, this task directly adapts the model on target videos and does not need re-training on both source and target data for each new domain.
Normally, the model (or segmentor) is trained with videos from the source domain $\mathcal{D}_{a}$ and tested on the target domain $\mathcal{D}_{b}$ where each video is provided with ground-truth mask of the first frame. By specifying the target in the first frame, the model can be used for the dense tracking tasks, e.g., multiple tool tracking, tool tip tracking, etc. We denote the target surgical video as $ \{(I_{1}, y_{1}), I_{2}, ..., I_{T}\} $, where $y_{1}$ is the given mask and $T$ is the length that varies among different videos. The first frame $(I_{1}, y_{1})$ is used for adaptation and the performance is evaluated on the predictions $\{\hat{y}_{2}, ..., \hat{y}_{T}\}$.

\begin{algorithm}
\caption{Video-level meta-training}
\label{alg:A}
\begin{algorithmic}[1]
\REQUIRE training video set: $\mathcal{D}_{a}$, learning rate $\beta_{\theta}, \beta_{\alpha}$
\STATE Initialize $\theta$, $\alpha$ \COMMENT{Initialize parameters}
\WHILE{\textit{not done}}
\STATE Sample $n$ videos \{$v_{1}$, $v_{2}$, ..., $v_{n}$\} from $\mathcal{D}_{a}$
\STATE $L\leftarrow0$         \COMMENT{Initialize accumulator}
\FOR{$k$ from 1 to $n$}
\STATE Sample a pair $\{(I_{i}, y_{i}), (I_{i+\varepsilon}, y_{i+\varepsilon})\}$ from $v_{k}$
\STATE $\theta^{\tau}\leftarrow \theta - \alpha\odot\nabla_{\theta}\mathcal{L}_{seg}( I_{i}, y_{i}; \theta)$ \COMMENT{meta-train}
\STATE $L \leftarrow L + \mathcal{L}_{seg}( I_{i+\varepsilon}, y_{i+\varepsilon}; \theta^{\tau})$ \COMMENT{meta-test}
\ENDFOR
\STATE $\theta^{*} \leftarrow \theta - \beta_{\theta} \nabla_{\theta}  \frac{1}{n}L$
\STATE $\alpha^{*}\leftarrow \alpha - \beta_{\alpha} \nabla_{\alpha} \frac{1}{n}L$ \COMMENT{Meta-learner}
\ENDWHILE
\ENSURE  $\theta^{*}, \alpha^{*}$
\end{algorithmic}
\end{algorithm}

\subsection{Video-level Meta-training}
We adopt a meta-learner ($\theta, \alpha$) to train the model $\mathcal{S}_{\theta}$ on the source dataset $D_{a}$ from parameters $\theta$ and step size $\alpha$ in order to teach the online adaptation during testing.
To achieve this, we learn the meta-learner to extract task-general knowledge through the experience of solving a number of related tasks.
Formally, we consider a family of tasks $\mathcal{T}$ that sampled from $D_{a}$. These tasks share some common structures such that learning to solve a single task has the potential to aid in solving another~\cite{grant2018recasting}. Each task $\tau\in \mathcal{T}$ defines a distribution over data points, which we assume in this work to consists of frames $I$ and either ground-truth masks $y$ in a supervised manner. It is further split into meta-train and meta-test sets, such that minimizing a task-specific performance metric (e.g., segmentation loss) corresponding to any unseen task given only a small number of data from the task, enables the fast adaptation to a new task via one or few steps of gradient descent.
In the context of video-level instrument segmentation,  we define the task $\tau $ by segmenting a pair of frames $(I_{i}, I_{i+\varepsilon})$. The first frame ($I_{i}, y_{i}$) and the subsequent frame ($I_{i+\varepsilon}, y_{i+\varepsilon}$) in the pair are used to compose the meta-train and meta-test sets for the task-specific evaluation. We adopt the randomly sampled subsequent frame  to simulate the possible variations in videos and enhance the generalization of the meta-learner.
%To achieve this, we learn the meta-learner that can capture the task-general knowledge by solving a set of similar video-specific tasks $\mathcal{T}$ sampled from  the source set $D_{a}$. Each task $\tau\in \mathcal{T}$ is associated with dataset $D_{\tau}$ that is further split into meta-train set $D_{\tau}^{tr}$ and meta-test set $D_{\tau}^{test}$.

%Learning to handle the task in $D_{\tau}^{tr}$ is helpful to tackle the task in $D_{\tau}^{test}$. The fast adaptation capability can  be obtained by optimizing the meta-train task using one step of gradient descent such that the meta-test task can be well adapted with only one update of  parameters.

%In the context of video-level instrument segmentation,  we define the task $\tau $ by segmenting a pair of frames $(I_{i}, I_{i+\varepsilon})$. The first frame $I_{i}$ and the subsequent frame $I_{i+\varepsilon}$ in the pair are used to meta-train and meta-test the model respectively, so that the updated parameters can be quickly adapted to an unseen video. We adopt the randomly sampled subsequent frame for meta-test to simulate the possible variations in videos and enhance the generalization of the meta-learner.

The full algorithm is outlined in Algorithm~\ref{alg:A} and consists of two For-loops. We firstly sample $n$ videos from $D_{a}$. The meta-train and meta-test sets are formulated by sampling the frame pair from each selected video $v_{k}$.
To quickly learn the task-specific parameters in the inner For-loop,
we update the model on meta-train set using one step of gradient descent:
\begin{equation}
\label{eq:1}
    \theta^{\tau}\leftarrow \theta - \alpha\odot\nabla_{\theta} \mathcal{L}_{seg}( I_{i}, y_{i} ; \theta),
\end{equation}
where the step size $\alpha$  has the same size as $\theta$ and $\odot$ denotes the element-wise product. Similar to the work~\cite{zhang2019online}, $\alpha$ is learnable instead of being fixed. $\mathcal{L}_{seg}$ is a segmentation loss that can be in different forms. Here, we use a hybrid loss that combines the cross-entropy term and the jaccard term, which is the same as~\cite{jin2019incorporating}.
The outer For-loop is the meta-learner that is optimized over the meta-test set, which can produce good task specific parameters after adaptation:
\begin{equation}
\label{eq:2}
\theta^{*} \leftarrow \theta - \beta_{\theta} \nabla_{\theta}  \frac{1}{n}L,~~
\alpha^{*}\leftarrow \alpha - \beta_{\alpha} \nabla_{\alpha} \frac{1}{n}L,
\end{equation}
where $L$ is the accumulated loss evaluated on $n$ videos, $\beta_{\theta}$ and $\beta_{\alpha}$ denote the fixed learning rates for updating $\theta$ and $\alpha$ in the meta-learner. Such learning scheme finally provides an optimal initialization ($\theta^{*}, \alpha^{*}$) that enables the model $\mathcal{S}_{\theta^{*}}$ to fast adapt to new videos from target domain $D_{b}$.
%Through such meta-learning, the trained model $\mathcal{S}_{\theta^{*}}$ can capture the common knowledge shared across different instrument segmentation tasks and fast adapt to the new videos with guaranteed performance.

\subsection{Dynamic Online Adaptation}
After finishing meta-training, we can use the meta-learner ($\theta^{*}, \alpha^{*}$) to initialize and update the model on the first frame mask ($I_{1}, y_{1}$) of the testing videos from the target dataset $D_{b}$ (\textbf{line 4} in Algorithm~\ref{alg:B}). Benefited from the general knowledge of instruments extracted by the meta-learner, the model can well fit the target instrument in the ($I_{1}, y_{1}$) with merely a few iterations $K$ ($K<=5$). In contrast, the normal trained model needs hundreds of fine-tuning iterations to fit the target, which is time-consuming during inference (see the comparison in Table~\ref{tab:result}).

Only learning from one shot of the target instruments is insufficient to deal with the large appearance variation in  the subsequent frames.
%change, motion blur, and occlusion that occur in the subsequent frames.
To avoid the segmentation performance decay in the long-range videos, previous works induce pseudo mask from confidence estimation~\cite{xiao2019online} or motion flows~\cite{zhao2020learning} across all testing frames to online update model parameters.
Unfortunately, the induced  masks show low quality when encountering the fast instrument motions in robotic surgical videos.
Constantly adapting the model on all testing frames with these masks  brings in the noisy supervision that easily leads to the target drift or missing after error accumulation.
%Unfortunately, both of them fail to handle the fast instrument motions between continuous frames in surgical videos, which may easily cause target drift or missing due to the  cumulative errors in noisy supervision.
To this regard, we propose to dynamically and selectively adapt the model across the testing frames with a noise-aware online adaptation strategy.
\begin{algorithm}
\caption{Dynamic online adaptation}
\label{alg:B}
\begin{algorithmic}[1]
\REQUIRE testing video set $D_{b}$, $(\theta^{*}, \alpha^{*})$, fine-tuning iteration $K$, balance weight $\gamma$
\STATE Testing $v_{j}=\{ (I_{1}, y_{1})), I_{2},.., I_{T}\}$ from $D_{b}$
\STATE $\theta^{1} \leftarrow \theta^{*}$, $\hat{y}_{1}\leftarrow y_{1}$
\FOR{$i$ from 1 to $K$}
\STATE $\theta^{1}\leftarrow \theta^{1} - \alpha^{*}\odot \nabla_{\theta^{1}}\mathcal{L}_{seg}(I_{1}, y_{1}; \theta^{1})$ \COMMENT{Meta-adaptation}
\ENDFOR
\FOR{$i$ from 2 to $T$}
\STATE $\bar{y}_{i} \leftarrow \texttt{generate\_pseudomask}(\hat{y}_{i-1}$)
\STATE $\theta^{i} \leftarrow \theta^{i-1}$
\STATE $\nabla_{\theta^{i}}\mathcal{L}\!\leftarrow\! \gamma\nabla_{\theta^{i}}\mathcal{L}_{seg}(I_{1}, y_{1} ;
\theta^{i}) \!+\! (1\!-\!\gamma)\nabla_{\theta^{i}}\mathcal{L}_{pse}(I_{i}, \bar{y}_{i}; \theta^{i})$
\IF {$ \texttt{gradient\_gate}(\nabla_{\theta^{i}}\mathcal{L})<0$}
\STATE $\theta^{i} \leftarrow \theta^{i} - \alpha^{*}
\odot\nabla_{\theta^{i}}\mathcal{L}$ \COMMENT{Online adaptation}
\ELSE
\STATE {$\theta^{i} \leftarrow \theta^{i}$} \COMMENT{No adaptation}
\ENDIF
\STATE $ \hat{y}_{i} \leftarrow \mathcal{S}_{\theta^{i}}(I_{i})$
\ENDFOR

\ENSURE Predictions $\{\hat{y}_{2},\hat{y}_{3},..., \hat{y}_{T} \}$
\end{algorithmic}
\end{algorithm}

\subsubsection{Pseudo mask generation}
%In the proposed meta-learning based paradigm, it is critical to estimate the gradient of the testing frames without corresponding ground truth.
We first generate the pseudo mask $\bar{y}_{i}$ for the current frame $I_{i}$ from the last frame prediction $\hat{y}_{i-1}$ and updated parameters $\theta^{i-1}$. Following~\cite{voigtlaender2017online}, we regard the pixels that are far from the predicted $\hat{y}_{i-1}$ as the background $\bar{y}_{i}^{b}$. It is obtained by thresholding the distance map with a pre-defined value $\tau_{b}$. The distance map is transformed by calculating the Euclidean distance to the closest foreground pixels in   $\hat{y}_{i-1}$. The high confidence elements (larger than $\tau_{f}$) in the prediction $\mathcal{S}_{\theta^{i-1}}(I_{i})$ is considered as the foreground $\bar{y}_{i}^{f}$. The regions that neither belong to $\bar{y}_{i}^{b}$ nor $\bar{y}_{i}^{f}$ are assigned an $ignore$ label.
%The undefined regions $\bar{y}_{i}^{u}$ that neither belong to foreground nor background are considered as hard pixels and assigned an $ignore$ label.
%The  noisy mask finally consists of $\bar{y}_{i}^{f}\cup \bar{y}_{i}^{b}\cup \bar{y}_{i}^{u}$.
Once obtaining the pseudo mask $\bar{y}_{i}$, we use a soft cross-entropy loss to estimate the gradient of the current frame:
\begin{equation}
  \label{eq:3}
  \mathcal{L}_{pse}= \frac{1}{HW}\sum_{t}-\bar{y}_{i,t}^{f}
  \mathrm{log}\hat{y}_{i,t} - \bar{y}_{i,t}^{b}
  \mathrm{log}(1-\hat{y}_{i,t}),
\end{equation}
where $t$ is the pixel index, $HW$ is the prediction size.  $\mathcal{L}_{pse}$ imposes supervision on pixels that are confident to the foreground or background, while keeping the model away from being disrupted by hard pixels.

\subsubsection{Dynamic adaptation with gradient gate}
In order to avoid the negative effect of the noisy supervision and increase the model robustness to tackle large temporal variation,
we propose to selectively update the model by leveraging the available label of first frame to assess the quality of each pseudo mask.
Only the clean pseudo samples that can minimize the discrepancy towards ground truth are used to guide the model adaptation.
%We form a gradient gate to conduct the quality assessment.
Specifically, we first move forward to explore the updated model after one gradient descent.
Apart from the gradient derived from the pseudo masks, we integrate the gradient calculated from the first frame with its ground truth to support the adaptation:
%The overall gradient is designed as follows:
\begin{equation}
\begin{aligned}
\label{eq:4}
\nabla_{\theta^{i}}\mathcal{L}&\leftarrow \gamma\nabla_{\theta^{i}}\mathcal{L}_{seg}(I_{1}, y_{1} ;
\theta^{i})\\&+ (1\!-\!\gamma)\nabla_{\theta^{i}}\mathcal{L}_{pse}(I_{i}, \bar{y}_{i}; \theta^{i}),
\end{aligned}
\end{equation}
where $\gamma$ is a hyper-parameter for balancing each term.
The overall gradient in Eq.~\eqref{eq:4} not only introduces information from previous adjacent frame but also provides the trusted update direction from ground truth.
\begin{figure}[t]
	\centering
	\includegraphics[width=0.40\textwidth]{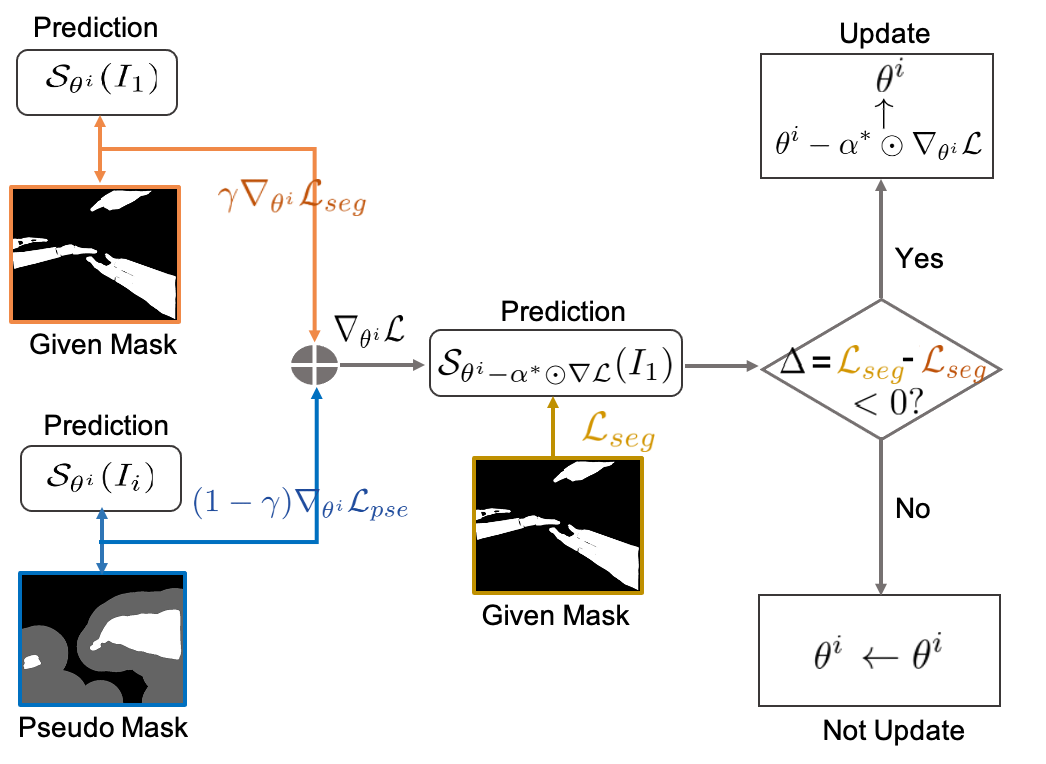}
	\caption{Gradient gate pipeline. (i) Each adaptation step starts with the combination of current frame gradient and first frame gradient. (ii) The overall gradient is used to re-evaluate the first frame and calculate $\Delta$ to filter out the imperfect supervision.}
	\label{gate}
	\vspace{-3mm}
\end{figure}
%Fig.~\ref{gate} shows the pipeline of the proposed gradient gate that dynamically selects the positive gradients for model adaptation and discards the negative gradients coming from the noisy supervision.
After gradient integration, we measure the loss difference on the first frame using the parameters after performing one gradient descent or not:
\begin{equation}
\label{eq:5}
\Delta=\mathcal{L}_{seg}(I_{1}, y_{1}; \theta_{i}-\alpha^{*}\odot\nabla_{\theta_{i}}\mathcal{L})-\mathcal{L}_{seg}(I_{1}, y_{1}; \theta_{i}),
\end{equation}
where $\Delta<0$ demonstrates that the induced pseudo mask has high quality, so that calculated gradient positively optimizes the model; while $\Delta>0$ means that the noise introduced by pseudo mask provides the adverse supervision on model update.
%where $\Delta<0$ demonstrates that the induced pseudo mask has high quality, so that the calculated gradient positively optimize the model on both the current and previous information, while $\Delta>0$ means that the noisy supervision has adverse effect on the previously learned knowledge.
We then form a gradient gate based on this loss measurement to dynamically adapt the model.
As shown in Fig.~\ref{gate},
%The proposed online adaptation dynamically adapt the model based on the output of the gradient gate.
if $\Delta<0$, the model will be updated using one gradient descent step:
\begin{equation}
\label{eq:6}
\theta^{i} \leftarrow \theta^{i} - \alpha^{*}
\odot\nabla_{\theta^{i}}\mathcal{L},
\end{equation}
otherwise, the model will not adapt on the current frame with no gradient back-propagated.
The proposed flexible online adaptation strategy can increase adaptation effectiveness to a large extent as the noisy supervision that mainly occurs in surgical videos with fast instrument motions can be mitigated.
Additionally, it enhances the adaptation efficiency with only a part proportion of testing frames requiring adaptation.
%The proposed flexible online adaptation strategy can increase both adaptation efficiency and effectiveness to a large extent because (1) only a part proportion of testing frames requires adaptation; (2) the noisy supervision that mainly occurs in surgical videos with fast instrument motions can be mitigated.

%======================================================
\section{EXPERIMENTS}
\label{EXPERIMENTS}
We validate our method on the binary instrument segmentation task. The segmentation model is trained with EndoVis17 (source domain) and then extensively evaluated on four different target domains.

\subsection{Datasets And Evaluation Metrics}

\textbf{Source}: We employ the EndoVis17, a public challenge dataset from 2017 MICCAI EndoVis Instrument Segmentation~\cite{allan20192017} for model training. It releases 8$\times$225-frame in-vivo surgical video sequences recorded from da Vinci Xi surgical system during different porcine procedures. We use all video data from the binary task for training.
%Each frame is of a high resolution of 1280$ \times $1024.

\textbf{Target}: We use two datasets for testing, including: (i) EndoVis18:  a larger surgical video dataset from 2018 MICCAI EndoVis Scene Segmentation Challenge~\cite{allan20202018}. It contains 15 in-vivo sequences in more complex surgery scenes acquired from  da Vinci X / Xi surgical system. For evaluation, we ignore other segmentation classes such as surgical  clips and anatomical objects. (ii) HKPWH: a newly collected robotic surgical dataset from Hong Kong Prince of Wales Hospital. It contains 8 fully annotated in-vivo sequences that record 4 real-world phases of prostatectomy performed by surgeons on da Vinci Xi surgical system. The videos are downsampled at 2 fps and contains 5 instruments. HKPWH captures the real-world RAS with surgeries performed on real patients. All videos in EndoVis18 and HKPWH are used for testing.

We employ two commonly used metrics including mean intersection-over-union (IoU) and Dice coefficient (Dice)  to quantitatively analyze the results, which are also widely used in previous works~\cite{jin2019incorporating,shvets2018automatic,zhao2020learning}.

%\subsection{Evaluation Metrics}
%We employ two commonly used metrics including mean intersection-over-union (IoU) and Dice coefficient (Dice)  to quantitatively analyze the results, which are also widely used in previous works~\cite{jin2019incorporating,shvets2018automatic,zhao2020learning}.
%We compute two metrics following $\mathrm{IoU}=\frac{|\mathrm{X} \cap \mathrm{Y}|}{|\mathrm{X}\cup\mathrm{Y}|}, \mathrm{Dice}=\frac{2|\mathrm{X} \cap \mathrm{Y}|}{|\mathrm{X}|+|\mathrm{Y}|}$, where $\mathrm{X}$ and $\mathrm{Y}$ denote the prediction set and ground-truth set respectively.

\subsection{Implementation Details}
%For each video, we provide the first frame annotation and the evaluation starts from the second frame, which is the same as the setup in video object segmentation task~\cite{xiao2019online}. The evaluation metrics are calculated on each video in the target dataset and then averaged. All the experiments are repeated 5 times to account for the randomness of DNN training.

The backbone of our segmentation model is U-Net11~\cite{ronneberger2015u} with pretrained encoders from VGG11~\cite{simonyan2014very}, which is the same as~\cite{jin2019incorporating, zhao2020learning}. As for meta-training, the meta-learner is trained with Adam optimizer and the batch size is set as 1. The learning rates $\beta_{\theta}$ and $\beta_{\alpha}$  are both set as $1e\!-\!5$ and multiplied by 0.1 after 10K iterations. The learnable step size $\alpha$ is initialized as $1e\!-\!4$. The sampling interval $\varepsilon$ varies in the range of [3,7]. For testing on EndoVis18 and HKPWH datasets, we meta-adapt the model on the first frame with $K=5$ iterations.
The other hyper-parameters for dynamic online adaptation are set as: $\gamma=0.5$, $\tau_{b}=90$,  $\tau_{f}=0.9$. The proposed method is implemented in PyTorch with a NVIDIA Titan Xp GPU.
All the experiments are repeated 5 times to account for the randomness of network training. The  metrics are first calculated on each video in the target dataset and then averaged to obtain the final results.

\begin{table}[t]
\caption{Segmentation Results of Different Methods on EndoVis18 and HKPWH datasets.}
\vspace{-1mm}
\label{tab:result}
\centering
\resizebox{85mm}{!}{
\begin{tabular}{c|cc|cccc}
\hline
\multirow{2}{*}{Method} &
  \multicolumn{2}{c|}{Testing} &
  \multicolumn{2}{c}{\begin{tabular}[c]{@{}c@{}}EndoVis17$\to$\\ EndoVis18\end{tabular}} &
  \multicolumn{2}{c}{\begin{tabular}[c]{@{}c@{}}EndoVis17$\to$\\ HKPWH\end{tabular}} \\ \cline{2-7}
 &
  \begin{tabular}[c]{@{}c@{}}FT\\ iter.\end{tabular} &
  \begin{tabular}[c]{@{}c@{}}FT\\ time\end{tabular} &
  \begin{tabular}[c]{@{}c@{}}IoU\\ (\%)\end{tabular} &
  \begin{tabular}[c]{@{}c@{}}Dice\\ (\%)\end{tabular} &
  \begin{tabular}[c]{@{}c@{}}IoU\\ (\%)\end{tabular} &
  \begin{tabular}[c]{@{}c@{}}Dice\\ (\%)\end{tabular} \\ \hline
Base            & 0   & 0s & 58.0 &  70.5       &    61.7    &     74.6 \\
Base-FT         & 100 & 25.16s & 72.4 & 82.5          & 76.3           & 85.5    \\
Base-FT         & 5   & 1.85s & 68.3 & 79.7          & 70.9           & 80.3   \\ \hline
%MVOS~\cite{xiao2019online}            & 5   & 0.18s & 71.4 & 82.3          & 72.0           & 82.4   \\
Dual-MF~\cite{zhao2020learning}          & 5   & 1.46s & 70.9 & 81.0          & 74.3           & 84.1  \\
MVOS-OL~\cite{xiao2019online}         & 5   & 1.67s & 71.0 & 81.9          & 74.7           & 84.6   \\\hline

Base-Meta & 5   & 1.46s & 72.2 & 82.5        & 73.8         & 83.4   \\
Base-Meta+online    & 5   & 1.46s & 71.7 & 81.1         & 75.5          & 84.9    \\
MDAL (Ours)        & 5   & 1.46s & \textbf{73.3}          & \textbf{83.6}          & \textbf{77.7}           & \textbf{86.2}   \\ \hline
\end{tabular}
}
\vspace{-3mm}
\end{table}
\subsection{Comparison with Other Methods}
We compare our method with several baselines, including the base model and fine-tuning baselines. For further comparisons with other methods, we adapt MVOS-OL~\cite{xiao2019online} and Dual-MF~\cite{zhao2020learning} to the video adaptation setting, so that they can be applied in our protocols. Our comparisons include: (i) \textit{Base}: the model trained on EndoVis17 and directly tested on target videos. (ii) \textit{Base-FT}: the model further fine-tuned on the first frame of target videos before testing. (iii) \textit{Dual-MF}~\cite{zhao2020learning}: the model meta-trained on EndoVis17 and adapted to target videos with online supervision induced from motion flows. (iv) \textit{MVOS-OL}~\cite{xiao2019online}: the model meta-trained on EndoVis17 and only specific layers adapted to target videos with online supervision induced from confidence estimation.  We use the same backbone (U-Net11) among these methods for fair comparison, except for the task-specific \textit{MVOS-OL}.

\begin{figure}[t]
	\centering
	\includegraphics[width=0.46\textwidth]{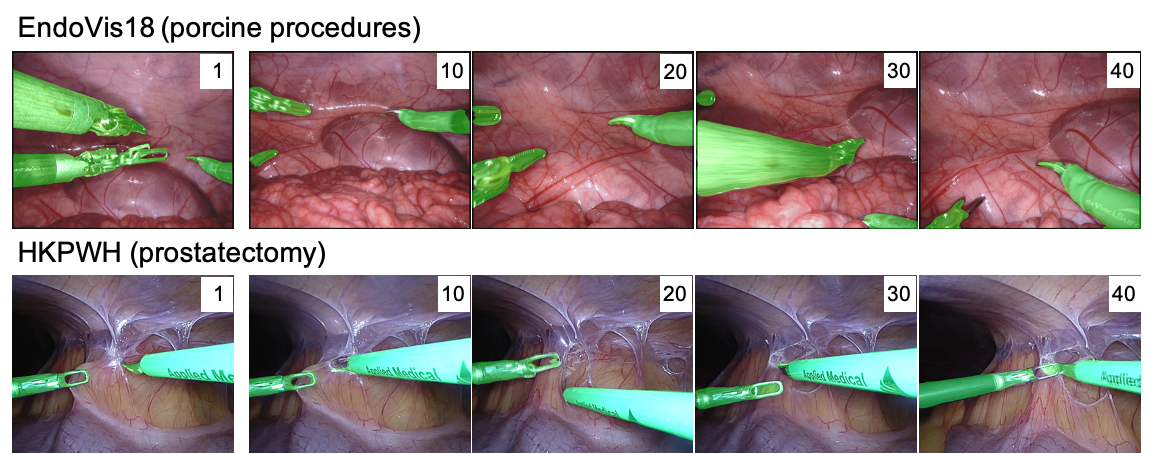}
	\caption{Qualitative results of proposed MDAL on EndoVis18 and HKPWH datasets. The first column shows the first frame annotation. }
	\label{vis}
	\vspace{-3mm}
\end{figure}

Table~\ref{tab:result} summarizes the adaptation results to EndoVis18 and HKPWH datasets. We report the fine-tuning (FT) time of each method, which counts a major proportion of the adaptation speed. The following observations can be made: (i) Overall, our method, requiring only 5 iterations of meta-adaptation on the first frame, achieves the best results on two datasets. It greatly shorten the FT time by 23.7s
compared with \textit{Base-FT} (100 iterations). (ii) The base model trained on EndoVis17 cannot generalize well on target domains, especially for EndoVis18. This is because of the distribution shift between two domains. (iii) Fine-tuning on the first frame can alleviate this problem but the adaptation speed slows down as the iteration number increases. %The performance increment is also limited because more iterations (over 100) bring little benefit.
%(iv) Meta-training on EndoVis17 (\textit{MVOS}) performs much better than \textit{Base-FT}, but is still not comparable to our method.
(iv) The performance of \textit{MVOS-OL} and
\textit{Dual-MF} is still not comparable to our method. We attribute it to the inaccurate online adaptation strategies that fail to handle the fast instrument motions in videos.
We further present some visual results of our method in Fig.~\ref{vis}. Some common challenges in robotic surgical videos, e.g., fast motion and multiple instances, can be alleviated by our method as it can still achieve robust and complete segmentation results.

\subsection{Ablation Studies}

\subsubsection{Effectiveness of key components}
We implement three configurations to analyze the effectiveness of key components:
(i) \textit{Base-Meta}: the model meta-trained on EndoVis17 and tested on target domain without online adaptation;
(ii) \textit{Base-Meta+online}: the model further tested on target domain  with naive online adaptation (i.e., no gradient gate);
(iii) \textit{MDAL}: the model meta-trained on EndoVis17 and tested on target domain  with dynamic online adaptation. In Table~\ref{tab:result}, the meta-learning baseline has already improved \textit{Base-FT} by a large margin. Further introducing naive online adaptation may degrade the performance instead, which is caused by the misleading supervision from the noisy pseudo masks. The dynamic online adaptation can reduce such issue to some degree, leading to consistent improvement, e.g., 2.5\% Dice on EndoVis18 and 1.3\% Dice on HKPWH.

 \begin{table}[t]
\caption{Adaptation Performance (mean IoU) on HKPWH under different fine-tuning iterations (K) on the first frame. }
\vspace{-1mm}
\label{tab:k}
\centering
\resizebox{83mm}{!}{
\begin{tabular}{c|ccccccc}
\hline
\multirow{2}{*}{Method} & \multicolumn{7}{c}{FT iterations (K) on the first frame} \\ \cline{2-8}
                        & K=0   & K=1   & K=3   & K=5   & K=7  & K=10  & K=100 \\ \hline
Base-FT                 & 61.7  & 64.5  & 67.3  & 70.9  & 71.6 & 73.1 & 76.3  \\
Base-Meta            & 67.1  & 69.8  & 71.2  & 73.8  & 75.5 & 76.0 & -     \\
MDAL                    & 73.7  & 75.8  & 76.5  & 77.7  & 77.9 & 78.3 & -     \\ \hline
\end{tabular}
}
\vspace{-3mm}
\end{table}

\subsubsection{Behavior of meta-learning}
Table~\ref{tab:k} shows the adaptation performance with different fine-tuning iterations ($K$) on the first frame. Without adaptation ($K=0$), our meta-trained model still achieves at least 5.4\% IoU higher than the fine-tuning baseline. This is mainly because that the proposed meta-learning scheme captures the general structure of foreground dynamic objects in the source domain such that the unseen target from a new distribution can be better recognized. For the baseline, the normal fine-tuning with several iterations brings limited improvement. It needs more iterations ($K=100$) to fit the target, especially when we do data augmentations on the first frame. In the contrary, by adopting the meta-learner ($\theta^{*}, \alpha^{*}$) to initialize the model, it can fast adapt to the new distribution with only several iterations. When $K$ increases to 5, our method has already outperformed \textit{Base-FT} with 100 iterations.
\begin{figure}[t]
	\centering
	\includegraphics[width=0.47\textwidth]{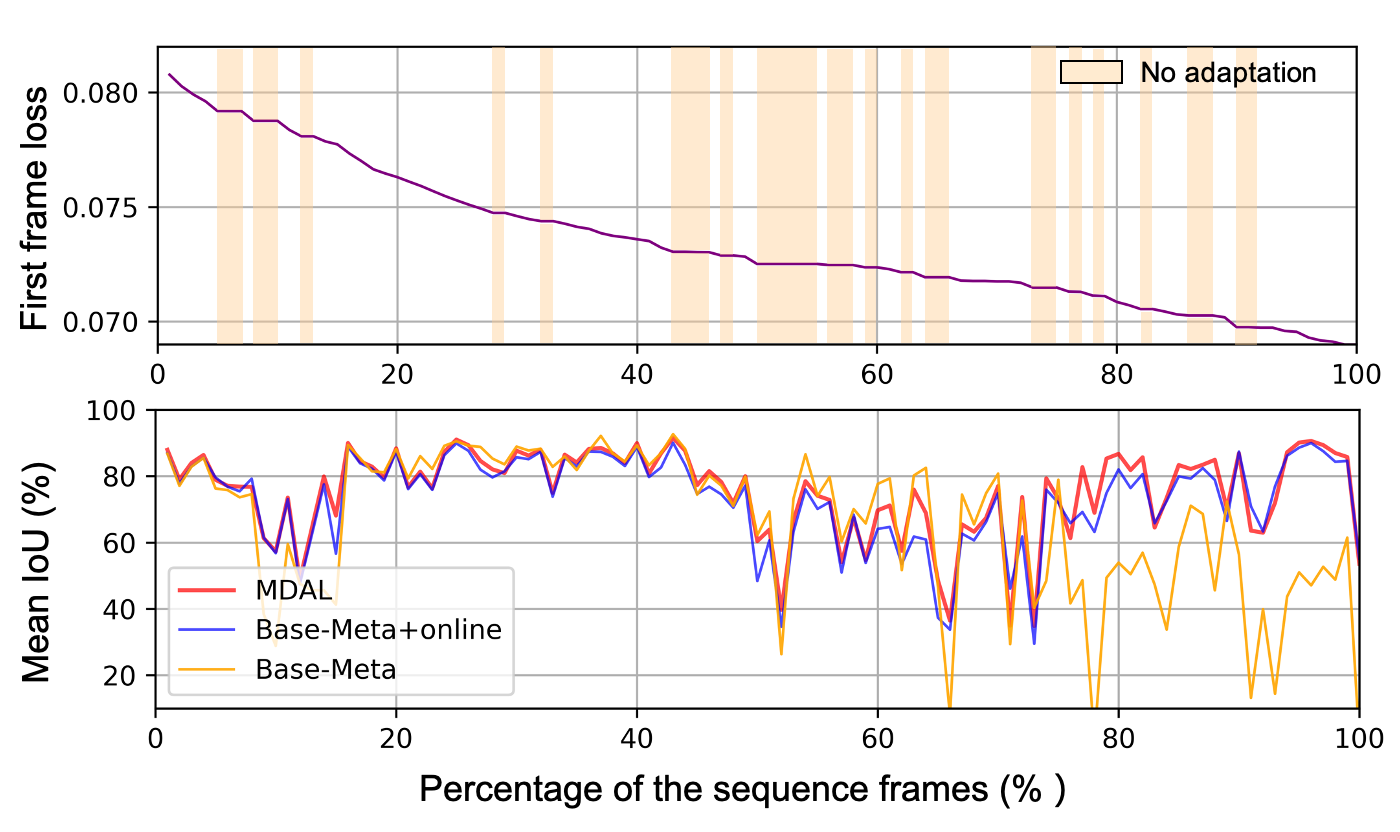}

	\caption{\textbf{Top row:} the process of dynamic online adaption in MDAL. Each adaptation step causes the decrease of the loss on the first frame. Yellow region denotes no adaptation. \textbf{Bottom row:} performance comparison (mean IoU) of different methods across frames on HKPWH (\texttt{seq\_1}).}
	\label{online}
	\vspace{-3mm}
\end{figure}
\subsubsection{Behavior of online adaptation} To explicitly show the process of proposed dynamic online adaptation, we visualize the first frame loss curve ($\mathcal{L}_{seg}$) over testing time in Fig.~\ref{online}.
It is observed that we only employ the clean pseudo labels that decline $\mathcal{L}_{seg}$ to update model, otherwise choose to suspend the current adaptation resulting in a constant $\mathcal{L}_{seg}$ shown in the yellow region.
%The yellow region where $\mathcal{L}_{seg}$ does not decline denotes a pause in our online adaptation.
Note that the x-axis is normalized by the percentage of the sequence length.
Fig.~\ref{online} also shows the performance comparison (mean IoU) along the temporal domain.
Without online adaptation, we find that the performance decreases over time due to the appearance variation between the first frame and subsequent frames. If we online adapt the model on following frames, the performance over the last 30\% frames can be largely improved. Notably, the proposed gradient gate contributes to a more stable performance than the naive online adaptation.

\subsection{Potential for Downstream Applications}
We further show the promising potential of our method on two crucial downstream applications for robotic surgery, including robot-assisted suturing and tool tip tracking for autonomous camera control. We directly adopt the model trained on EndoVis17 and adapt it to two ex-vivo robotic surgical videos with $K=10$ considering the larger distribution shift between source and target domains.

\subsubsection{Robot-assisted suturing}
\begin{figure}[t]
	%\vspace{-2mm}
	%\setlength{\abovecaptionskip}{0.1cm}
	%\setlength{\belowcaptionskip}{-0.3cm}
	\centering
	\includegraphics[width=0.48\textwidth]{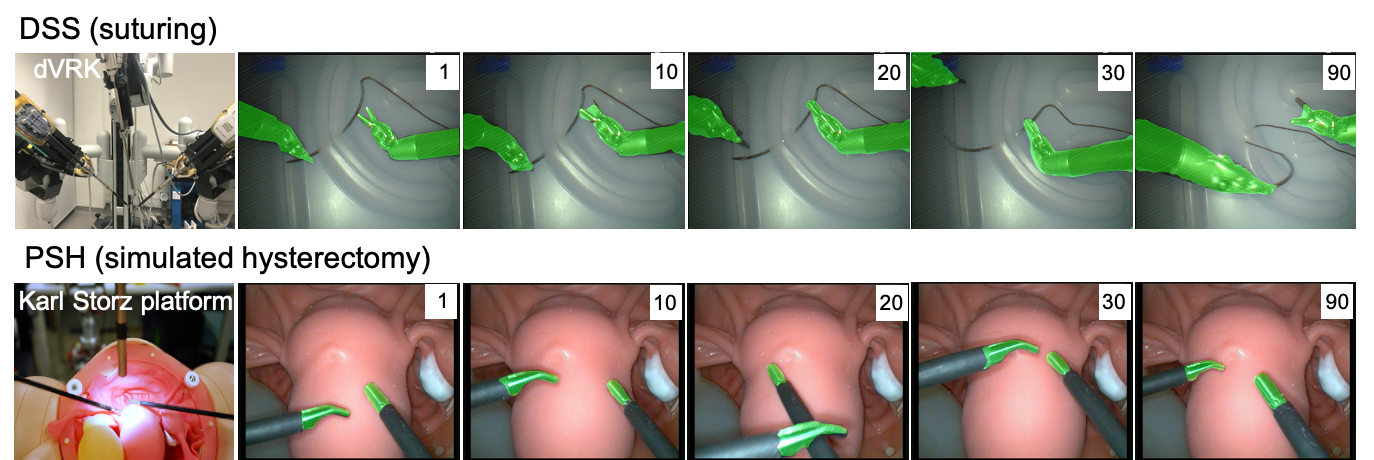}
	\caption{Qualitative results of proposed MDAL on DSS and PSH datasets. }
	\label{vis2}
	\vspace{-4mm}
\end{figure}

We record a robotic suturing procedure by da Vinci Research Kit (dVRK) system.
The collected dVRK-based surgical suturing (DSS) video is downsampled at 5 fps, which results in 158 frames for evaluation.
In Fig.~\ref{vis2}, we see that our method can well adapt the model to a substantially different scene, and consistently segment the robotic suturing instruments across frames.
Such segmentation results serve as the crucial shape information for image-guided automated suturing manipulations.
%\subsubsection{Tool tip tracking}
\subsubsection{Autonomous camera control} We adapt the model to another scene with a new task, i.e., dense tool tip tracking, for autonomous camera control. Specifically,
%We also explore the capability of our method to dense track the robotic tool tips across frames.
a phantom-based simulated hysterectomy (PSH) video is collected using Karl Storz Laparoscope and tools, and has 500 frames (at 8 fps) for evaluation.
%The video is downsampled at 8 fps and has 500 frames for evaluation.
Only adapting on the first frame with 2.6s, our model can continuously track the tiny tips of Karl Storz instruments used in robotic system (see Fig.~\ref{vis2}).
%Provided with the 3D tool position, the stereo camera can be automatically guided with optimal view in different surgeries.
%In this regard, our method shows the great potential for autonomous camera control with optimal view.
%The distribution shift between these two domains and source domain becomes larger compared with previous target domains.
%Our method still achieves successful adaptation with complete and consistent segmentation results, demonstrating the outstanding general efficacy of our method.
Provided with the tool position, the camera can be automatically adjusted for presenting the optimal view in surgery.
%where the optimal view is defined by the statistical heatmap in various procedures.

%======================================================
\section{CONCLUSIONS AND FUTURE WORK}
\label{CONCLUSIONS}

In this work, we propose a novel meta-learning based method for a new task, adaptive instrument segmentation from robotic surgical videos, given extremely scarce labels from target domain.
Our method addresses the difficulties in this new task (most notably distribution mismatch, followed by fast instrument motion and large appearance variation) by fully leveraging the meta-learning episodic training strategy targeted at video segmentation, as well as applying dynamic online adaptation across frames. It  achieves decent segmentation performance on different  RAS datasets by fast adapting one source model,
which exerts a significant impact in applicable research and clinical scenarios.
Although the current inference speed with 4 fps can demonstrate great potential for several essential robotic tasks, we shall explore how to increase the inference speed in future for wider application in robotic automation. We plan to incorporate a lightweight network~\cite{islam2019real, islam2019learning} into our learning scheme, or learn to update part of parameters for efficient adaptations.

%To further apply it to more fine-grained robotic tasks, the network architecture and learning rules can be optimized for a real-time analysis of target robotic surgical videos (currently 4.2 fps, good support for some downstream tasks). We plan to incorporate a lightweight network~\cite{islam2019real, islam2019learning} into our learning scheme, or learn to update part of parameters for efficient adaptations.

%\addtolength{\textheight}{-12cm}   % This command serves to balance the column lengths
	% on the last page of the document manually. It shortens
	% the textheight of the last page by a suitable amount.
	% This command does not take effect until the next page
	% so it should come on the page before the last. Make
	% sure that you do not shorten the textheight too much.

%\section*{ACKNOWLEDGMENT}

%The authors would like to thank Colin Lea for sharing raw features of JIGSAWS dataset.

\newpage

\bibliographystyle{IEEEtran}
\bibliography{IEEEabrv,mybibfile}

% Generated by IEEEtran.bst, version: 1.14 (2015/08/26)
\begin{thebibliography}{10}
\providecommand{\url}[1]{#1}
\csname url@samestyle\endcsname
\providecommand{\newblock}{\relax}
\providecommand{\bibinfo}[2]{#2}
\providecommand{\BIBentrySTDinterwordspacing}{\spaceskip=0pt\relax}
\providecommand{\BIBentryALTinterwordstretchfactor}{4}
\providecommand{\BIBentryALTinterwordspacing}{\spaceskip=\fontdimen2\font plus
\BIBentryALTinterwordstretchfactor\fontdimen3\font minus
  \fontdimen4\font\relax}
\providecommand{\BIBforeignlanguage}[2]{{%
\expandafter\ifx\csname l@#1\endcsname\relax
\typeout{** WARNING: IEEEtran.bst: No hyphenation pattern has been}%
\typeout{** loaded for the language `#1'. Using the pattern for}%
\typeout{** the default language instead.}%
\else
\language=\csname l@#1\endcsname
\fi
#2}}
\providecommand{\BIBdecl}{\relax}
\BIBdecl

\bibitem{allan20183}
M.~Allan, S.~Ourselin, D.~J. Hawkes, J.~D. Kelly, and D.~Stoyanov, ``3-d pose
  estimation of articulated instruments in robotic minimally invasive
  surgery,'' \emph{IEEE transactions on medical imaging}, vol.~37, no.~5, pp.
  1204--1213, 2018.

\bibitem{du2019patch}
X.~Du, M.~Allan, S.~Bodenstedt, L.~Maier-Hein, S.~Speidel, A.~Dore, and
  D.~Stoyanov, ``Patch-based adaptive weighting with segmentation and scale
  (pawss) for visual tracking in surgical video,'' \emph{Medical image
  analysis}, vol.~57, pp. 120--135, 2019.

\bibitem{allan20192017}
M.~Allan, A.~Shvets, T.~Kurmann, Z.~Zhang, R.~Duggal, Y.-H. Su, N.~Rieke,
  I.~Laina, N.~Kalavakonda, S.~Bodenstedt \emph{et~al.}, ``2017 robotic
  instrument segmentation challenge,'' \emph{arXiv preprint arXiv:1902.06426},
  2019.

\bibitem{gao2021future}
X.~Gao, Y.~Jin, Z.~Zhao, Q.~Dou, and P.-A. Heng, ``Future frame prediction for
  robot-assisted surgery,'' \emph{arXiv preprint arXiv:2103.10308}, 2021.

\bibitem{islam2020st}
M.~Islam, V.~Vibashan, C.~M. Lim, and H.~Ren, ``St-mtl: Spatio-temporal
  multitask learning model to predict scanpath while tracking instruments in
  robotic surgery,'' \emph{Medical Image Analysis}, p. 101837, 2020.

\bibitem{reiley2009task}
C.~E. Reiley and G.~D. Hager, ``Task versus subtask surgical skill evaluation
  of robotic minimally invasive surgery,'' in \emph{International conference on
  medical image computing and computer-assisted intervention}.\hskip 1em plus
  0.5em minus 0.4em\relax Springer, 2009, pp. 435--442.

\bibitem{zia2018surgical}
A.~Zia, A.~Hung, I.~Essa, and A.~Jarc, ``Surgical activity recognition in
  robot-assisted radical prostatectomy using deep learning,'' in
  \emph{International Conference on Medical Image Computing and
  Computer-Assisted Intervention}.\hskip 1em plus 0.5em minus 0.4em\relax
  Springer, 2018, pp. 273--280.

\bibitem{yu2019holistically}
L.~Yu, P.~Wang, X.~Yu, Y.~Yan, and Y.~Xia, ``A holistically-nested u-net:
  Surgical instrument segmentation based on convolutional neural network,''
  \emph{Journal of Digital Imaging}, pp. 1--7, 2019.

\bibitem{islam2019real}
M.~Islam, D.~A. Atputharuban, R.~Ramesh, and H.~Ren, ``Real-time instrument
  segmentation in robotic surgery using auxiliary supervised deep adversarial
  learning,'' \emph{IEEE Robotics and Automation Letters}, vol.~4, no.~2, pp.
  2188--2195, 2019.

\bibitem{islam2020ap}
M.~Islam, V.~VS, and H.~Ren, ``Ap-mtl: Attention pruned multi-task learning
  model for real-time instrument detection and segmentation in robot-assisted
  surgery,'' \emph{ICRA}, 2020.

\bibitem{ni2020attention}
Z.-L. Ni, G.-B. Bian, Z.-G. Hou, X.-H. Zhou, X.-L. Xie, and Z.~Li,
  ``Attention-guided lightweight network for real-time segmentation of robotic
  surgical instruments,'' in \emph{2020 IEEE International Conference on
  Robotics and Automation (ICRA)}.\hskip 1em plus 0.5em minus 0.4em\relax IEEE,
  2020, pp. 9939--9945.

\bibitem{jin2019incorporating}
Y.~Jin, K.~Cheng, Q.~Dou, and P.-A. Heng, ``Incorporating temporal prior from
  motion flow for instrument segmentation in minimally invasive surgery
  video,'' in \emph{International Conference on Medical Image Computing and
  Computer-Assisted Intervention}.\hskip 1em plus 0.5em minus 0.4em\relax
  Springer, 2019, pp. 440--448.

\bibitem{zhao2020learning}
Z.~Zhao, Y.~Jin, X.~Gao, Q.~Dou, and P.-A. Heng, ``Learning motion flows for
  semi-supervised instrument segmentation from robotic surgical video,'' in
  \emph{International Conference on Medical Image Computing and
  Computer-Assisted Intervention}.\hskip 1em plus 0.5em minus 0.4em\relax
  Springer, 2020, pp. 679--689.

\bibitem{da2019self}
C.~da~Costa~Rocha, N.~Padoy, and B.~Rosa, ``Self-supervised surgical tool
  segmentation using kinematic information,'' in \emph{2019 International
  Conference on Robotics and Automation (ICRA)}.\hskip 1em plus 0.5em minus
  0.4em\relax IEEE, 2019, pp. 8720--8726.

\bibitem{colleoni2020synthetic}
E.~Colleoni, P.~Edwards, and D.~Stoyanov, ``Synthetic and real inputs for tool
  segmentation in robotic surgery,'' in \emph{International Conference on
  Medical Image Computing and Computer-Assisted Intervention}.\hskip 1em plus
  0.5em minus 0.4em\relax Springer, 2020, pp. 700--710.

\bibitem{pfeiffer2019generating}
M.~Pfeiffer, I.~Funke, M.~R. Robu, S.~Bodenstedt, L.~Strenger, S.~Engelhardt,
  T.~Ro{\ss}, M.~J. Clarkson, K.~Gurusamy, B.~R. Davidson \emph{et~al.},
  ``Generating large labeled data sets for laparoscopic image processing tasks
  using unpaired image-to-image translation,'' in \emph{International
  Conference on Medical Image Computing and Computer-Assisted
  Intervention}.\hskip 1em plus 0.5em minus 0.4em\relax Springer, 2019, pp.
  119--127.

\bibitem{sahu2020endo}
M.~Sahu, R.~Str{\"o}msd{\"o}rfer, A.~Mukhopadhyay, and S.~Zachow,
  ``Endo-sim2real: Consistency learning-based domain adaptation for instrument
  segmentation,'' in \emph{International Conference on Medical Image Computing
  and Computer-Assisted Intervention}.\hskip 1em plus 0.5em minus 0.4em\relax
  Springer, 2020, pp. 784--794.

\bibitem{chen2019blending}
Z.~Chen, J.~Zhuang, X.~Liang, and L.~Lin, ``Blending-target domain adaptation
  by adversarial meta-adaptation networks,'' in \emph{Proceedings of the IEEE
  conference on computer vision and pattern recognition}, 2019, pp. 2248--2257.

\bibitem{qian2019domain}
K.~Qian and Z.~Yu, ``Domain adaptive dialog generation via meta learning,'' in
  \emph{Proceedings of the 57th Annual Meeting of the Association for
  Computational Linguistics}, 2019, pp. 2639--2649.

\bibitem{finn2017model}
C.~Finn, P.~Abbeel, and S.~Levine, ``Model-agnostic meta-learning for fast
  adaptation of deep networks,'' in \emph{ICML}, 2017.

\bibitem{sun2019meta}
Q.~Sun, Y.~Liu, T.-S. Chua, and B.~Schiele, ``Meta-transfer learning for
  few-shot learning,'' in \emph{Proceedings of the IEEE conference on computer
  vision and pattern recognition}, 2019, pp. 403--412.

\bibitem{jamal2019task}
M.~A. Jamal and G.-J. Qi, ``Task agnostic meta-learning for few-shot
  learning,'' in \emph{Proceedings of the IEEE conference on computer vision
  and pattern recognition}, 2019, pp. 11\,719--11\,727.

\bibitem{garcia2017few}
V.~Garcia and J.~Bruna, ``Few-shot learning with graph neural networks,''
  \emph{arXiv preprint arXiv:1711.04043}, 2017.

\bibitem{zhang2019online}
Z.~Zhang, S.~Lathuili{\`e}re, A.~Pilzer, N.~Sebe, E.~Ricci, and J.~Yang,
  ``Online adaptation through meta-learning for stereo depth estimation,''
  \emph{arXiv preprint arXiv:1904.08462}, 2019.

\bibitem{tonioni2019learning}
A.~Tonioni, O.~Rahnama, T.~Joy, L.~D. Stefano, T.~Ajanthan, and P.~H. Torr,
  ``Learning to adapt for stereo,'' in \emph{Proceedings of the IEEE conference
  on computer vision and pattern recognition}, 2019, pp. 9661--9670.

\bibitem{yu2020foal}
H.~Yu, S.~Sun, H.~Yu, X.~Chen, H.~Shi, T.~S. Huang, and T.~Chen, ``Foal: Fast
  online adaptive learning for cardiac motion estimation,'' in
  \emph{Proceedings of the IEEE conference on computer vision and pattern
  recognition}, 2020, pp. 4313--4323.

\bibitem{xiao2019online}
H.~Xiao, B.~Kang, Y.~Liu, M.~Zhang, and J.~Feng, ``Online meta adaptation for
  fast video object segmentation,'' \emph{IEEE transactions on pattern analysis
  and machine intelligence}, vol.~42, no.~5, pp. 1205--1217, 2019.

\bibitem{allan20202018}
M.~Allan, S.~Kondo, S.~Bodenstedt, S.~Leger, R.~Kadkhodamohammadi, I.~Luengo,
  F.~Fuentes, E.~Flouty, A.~Mohammed, M.~Pedersen \emph{et~al.}, ``2018 robotic
  scene segmentation challenge,'' \emph{arXiv preprint arXiv:2001.11190}, 2020.

\bibitem{grant2018recasting}
E.~Grant, C.~Finn, S.~Levine, T.~Darrell, and T.~Griffiths, ``Recasting
  gradient-based meta-learning as hierarchical bayes,'' in \emph{6th
  International Conference on Learning Representations, ICLR 2018}, 2018.

\bibitem{voigtlaender2017online}
P.~Voigtlaender and B.~Leibe, ``Online adaptation of convolutional neural
  networks for video object segmentation,'' \emph{arXiv preprint
  arXiv:1706.09364}, 2017.

\bibitem{shvets2018automatic}
A.~A. Shvets, A.~Rakhlin, A.~A. Kalinin, and V.~I. Iglovikov, ``Automatic
  instrument segmentation in robot-assisted surgery using deep learning,'' in
  \emph{2018 17th IEEE International Conference on Machine Learning and
  Applications (ICMLA)}.\hskip 1em plus 0.5em minus 0.4em\relax IEEE, 2018, pp.
  624--628.

\bibitem{ronneberger2015u}
O.~Ronneberger, P.~Fischer, and T.~Brox, ``U-net: Convolutional networks for
  biomedical image segmentation,'' in \emph{International Conference on Medical
  image computing and computer-assisted intervention}.\hskip 1em plus 0.5em
  minus 0.4em\relax Springer, 2015, pp. 234--241.

\bibitem{simonyan2014very}
K.~Simonyan and A.~Zisserman, ``Very deep convolutional networks for
  large-scale image recognition,'' \emph{arXiv preprint arXiv:1409.1556}, 2014.

\bibitem{islam2019learning}
M.~Islam, Y.~Li, and H.~Ren, ``Learning where to look while tracking
  instruments in robot-assisted surgery,'' in \emph{International Conference on
  Medical Image Computing and Computer-Assisted Intervention}.\hskip 1em plus
  0.5em minus 0.4em\relax Springer, 2019, pp. 412--420.

\end{thebibliography}

\end{document}